\documentclass[11pt,a4paper]{article}
\usepackage{acl2015}
\usepackage{times}
\usepackage{latexsym}

\usepackage{url}
\usepackage[leqno, fleqn]{amsmath}
\usepackage{amssymb}
\usepackage{qtree}
\usepackage{graphicx}
\usepackage{booktabs}
\usepackage{multirow}
\usepackage{colortbl}
\usepackage{caption}
\usepackage{subcaption}
\usepackage{color}
\usepackage{xcolor}
\usepackage{tikz}
\usepackage{ifthen}
\usepackage{framed}

\def\unknown{\textit{unknown}}
\def\contradiction{\textit{contradiction}}
\def\hypothesis{hypothesis}
\def\premise{premise}

\usepackage{color}

\newcommand\Fig[4]{\begin{figure}[tb] \begin{center} \includegraphics[scale=#2]{#1} \end{center} \caption{\label{fig:#3} #4} \end{figure}}

\newcommand\Reffig[1]{Figure~\ref{#1}}

\newcommand\reffig[1]{Figure~\ref{#1}}

\newcommand\reftab[1]{Table~\ref{#1}}
\newcommand\Reftab[1]{Table~\ref{#1}}
\newcommand\reftabs[2]{Tables~\ref{#1} and \ref{#2}}

\newcount\colveccount
\newcommand*\colvec[1]{
        \global\colveccount#1
        \begin{bmatrix}
        \colvecnext
}
\def\colvecnext#1{
        #1
        \global\advance\colveccount-1
        \ifnum\colveccount>0
                \\
                \expandafter\colvecnext
        \else
                \end{bmatrix}
        \fi
}

\newlength{\howlong}

\usepackage{stmaryrd}

\usepackage{gb4e}
\noautomath

\def\ii#1{\textit{#1}}
\newcommand{\word}[1]{\emph{#1}}
\newcommand{\fulllabel}[2]{\b{#1}\newline\textsc{#2}}

\makeatletter
\def\citealt{\def\citename##1{{\frenchspacing##1} }\@internalcitec}
\def\@citexc[#1]#2{\if@filesw\immediate\write\@auxout{\string\citation{#2}}\fi
  \def\@citea{}\@citealt{\@for\@citeb:=#2\do
    {\@citea\def\@citea{;\penalty\@m\ }\@ifundefined
       {b@\@citeb}{{\bf ?}\@warning
       {Citation `\@citeb' on page \thepage \space undefined}}%
{\csname b@\@citeb\endcsname}}}{#1}}
\def\@internalcitec{\@ifnextchar [{\@tempswatrue\@citexc}{\@tempswafalse\@citexc[]}}
\def\@citealt#1#2{{#1\if@tempswa, #2\fi}}
\makeatother

\title{A large annotated corpus for learning natural language inference}

\author{
Samuel R.\ Bowman$^{\ast\dag}$ \\
\texttt{sbowman@stanford.edu} \\
\And
Gabor Angeli$^{\dag\ddag}$ \\
\texttt{angeli@stanford.edu} \\
\AND
Christopher Potts$^{\ast}$\\
\texttt{cgpotts@stanford.edu}
\And
Christopher D.\ Manning$^{\ast\dag\ddag}$\\
\texttt{manning@stanford.edu}\\
\AND\\[-3ex]
{$^{\ast}$Stanford Linguistics\quad
$^{\dag}$Stanford NLP Group\quad
$^{\ddag}$Stanford Computer Science}
}

\date{}

\makeatletter
\newcommand{\@BIBLABEL}{\@emptybiblabel}
\newcommand{\@emptybiblabel}[1]{}
\definecolor{black}{rgb}{0,0,0}
\makeatother
\usepackage[breaklinks, colorlinks, linkcolor=black, urlcolor=black, citecolor=black]{hyperref}

\def\t#1{#1}
\def\b#1{\t{\textbf{#1}}}
\def\colspaceS{2.25mm}

\def\colspaceL{4.25mm}

\begin{document}
\maketitle

\begin{abstract} 
Understanding entailment and contradiction is  fundamental to understanding natural language, and inference about entailment and contradiction is a valuable testing ground for the development of semantic representations. However, machine learning research in this area has been dramatically limited by the lack of large-scale resources. To address this, we introduce the Stanford Natural Language Inference corpus, a new, freely available collection of labeled sentence pairs, written by humans doing a novel grounded task based on image captioning. At 570K pairs, it is two orders of magnitude larger than all other resources of its type. This increase in scale allows lexicalized classifiers to outperform some sophisticated existing entailment models, and it allows a neural network-based model to perform competitively on natural language inference benchmarks for the first time.
\end{abstract}

\begin{table*}[t]
  \centering\small
  \begin{tabular}{p{6.4cm}p{1.7cm}p{6.4cm}}
  \toprule
A man inspects the uniform of a figure in some East Asian country. & \fulllabel{contradiction}{c c c c c} & The man is sleeping\\
\rule{0pt}{3ex}An older and younger man smiling. & \fulllabel{neutral}{n n e n n} & Two men are smiling and laughing at the cats playing on the floor.\\
\rule{0pt}{3ex}A black race car starts up in front of a crowd of people. & \fulllabel{contradiction}{c c c c c} & A man is driving down a lonely road.\\
\rule{0pt}{3ex}A soccer game with multiple males playing. & \fulllabel{entailment}{e e e e e} & Some men are playing a sport.\\
\rule{0pt}{3ex}A smiling costumed woman is holding an umbrella. & \fulllabel{neutral}{n n e c n} & A happy woman in a fairy costume holds an umbrella.\\
    \bottomrule
  \end{tabular}
  \caption{\label{snli-examples}Randomly chosen examples from the development section of our new corpus, shown with both the selected gold labels and the full set of labels (abbreviated) from the individual annotators, including (in the first position) the label used by the initial author of the pair.}
\end{table*}

\section{Introduction}\label{sec:introduction}

The semantic concepts of entailment and contradiction are central to
all aspects of natural language meaning
\cite{Katz72,vanBenthem08NATLOG}, from the lexicon to the content of
entire texts. Thus, \emph{natural language
  inference} (NLI) --- characterizing and using these relations in
computational systems
\cite{Fyodorov-etal:2000,Condoravdi-etAl:2003,BosMar:2005,dagan2006pascal,maccartney2009extended} --- is
essential in tasks ranging from information retrieval to semantic
parsing to commonsense reasoning.

NLI has been addressed using a variety of techniques, including
those based on symbolic logic, knowledge bases, and neural networks. 
In recent years, it has become an important testing ground for
approaches employing \emph{distributed} word and phrase
representations. Distributed representations excel at capturing
relations based in similarity, and have proven effective at
modeling simple dimensions of meaning like evaluative sentiment
(e.g., \citealt{socher2013recursive}), but it is less clear that they can be
trained to support the full range of logical and commonsense
inferences required for NLI \cite{Bowman:Potts:Manning:2014,Weston-etal:2015,Weston-etal:2015Q}. 
In a SemEval 2014 task aimed at evaluating distributed
representations for NLI, the best-performing systems relied heavily on
additional features and reasoning capabilities
\cite{marelli2014semeval}. 

Our ultimate objective is to provide an empirical
evaluation of learning-centered approaches to NLI,
advancing the case for NLI as a tool for the evaluation of 
domain-general approaches to semantic representation. 
However, in our view, existing NLI corpora do not
permit such an assessment. They are generally too small for training
modern data-intensive, wide-coverage models, many contain sentences
that were algorithmically generated, and they are often beset with
indeterminacies of event and entity coreference that significantly
impact annotation quality.

To address this, this paper introduces the Stanford Natural Language
Inference (SNLI) corpus, a collection of sentence pairs labeled for
entailment, contradiction, and semantic independence. At 570,152
sentence pairs, SNLI is two orders of magnitude larger than all
other resources of its type. And, in contrast to many such resources,
all of its sentences and labels were written by humans in a grounded,
naturalistic context. In a separate validation phase, we collected
four additional judgments for each label for 56,941 of the examples.
Of these, 98\% of cases emerge with a three-annotator consensus, 
and 58\% see a unanimous consensus from all five annotators.

In this paper, we use this corpus to evaluate a variety of models
for natural language inference, including rule-based systems, simple
linear classifiers, and neural network-based models. 
We find that two models achieve comparable performance: a feature-rich
classifier model and a neural network model centered around a Long Short-Term Memory network (LSTM; 
\citealt{hochreiter1997long}). We further evaluate the LSTM model
by taking advantage of its ready support for transfer learning, and show that it can be adapted to an existing
NLI challenge task, yielding the best reported performance by a neural network model and approaching the overall state of the art.

\section{A new corpus for NLI}\label{sec:discussion}

To date, the primary sources of annotated NLI corpora have been the
Recognizing Textual Entailment (RTE)
challenge tasks.\footnote{\url{http://aclweb.org/aclwiki/index.php?title=Textual_Entailment_Resource_Pool}}
These are generally high-quality, hand-labeled data sets, and they
have stimulated innovative logical and statistical models of natural
language reasoning, but their small size (fewer than a thousand examples each)
limits their utility as a testbed for learned distributed representations. 
The data for the SemEval 2014 task called Sentences Involving Compositional Knowledge (SICK) is a step up in terms of size, but only to 4,500 training examples, and its
partly automatic construction introduced some spurious patterns into
the data (\citealt{marelli2014semeval}, $\S$6). The
Denotation Graph entailment set \cite{hodoshimage} contains millions of
examples of entailments between sentences and artificially constructed
short phrases, but it was labeled using fully automatic methods, and is
noisy enough that it is probably suitable only as a source of
supplementary training data.
Outside the domain of sentence-level entailment, \newcite{levy2014focused} introduce
a large corpus of semi-automatically annotated entailment examples 
between subject--verb--object relation triples, and the second release of the 
Paraphrase Database \cite{ganitkevitch2ppdb} includes automatically generated entailment annotations
over a large corpus of pairs of words and short phrases.

Existing resources suffer from a subtler issue that impacts even
projects using only human-provided annotations: indeterminacies of
event and entity coreference lead to insurmountable indeterminacy
concerning the correct semantic label (\citealt{de2008finding} $\S4.3$; \citealt{marelli2014sick}). 
For an example of the pitfalls surrounding entity coreference, consider the sentence pair \word{A boat sank in the Pacific Ocean} and \word{A boat sank in the Atlantic Ocean}. The pair could be labeled as a contradiction if one assumes that the two sentences refer to the same single event, but could also be reasonably labeled as neutral if that assumption is not made. In order to ensure that our labeling scheme assigns a single correct label to every pair, we must select one of these approaches across the board, but both choices present problems. If we opt not to assume that events are coreferent, then we will only ever find contradictions between sentences that make broad universal assertions, but if we opt to assume coreference, new counterintuitive predictions emerge. For example, \ii{Ruth Bader Ginsburg was appointed to the US Supreme Court} and \ii{I had a sandwich for lunch today} would unintuitively be labeled as a contradiction, rather than neutral, under this assumption.

Entity coreference presents a similar kind of indeterminacy, as in the pair \word{A tourist visited New York} and \word{A tourist visited the city}. Assuming coreference between \ii{New York} and \ii{the city} justifies labeling the pair as an entailment, but without that assumption \ii{the city} could be taken to refer to a specific unknown city, leaving the pair neutral. This kind of indeterminacy of label can be resolved only once the questions of coreference are resolved.

With SNLI, we sought to address the issues of size, quality, and
indeterminacy. To do this, we employed a crowdsourcing framework with
the following crucial innovations. First, the examples were grounded
in specific scenarios, and the premise and hypothesis sentences in each example 
were constrained to describe that scenario from the same perspective, 
which helps greatly in controlling event and entity coreference.\footnote{
Issues of coreference are not completely solved, but greatly mitigated. For example, with the premise sentence \word{A dog is lying in the grass}, a worker could safely assume that the dog is the most prominent thing in the photo, and very likely the only dog, and  build contradicting sentences assuming reference to the same dog.
} 
Second, the prompt
gave participants the freedom to produce entirely novel sentences
within the task setting, which led to richer examples than we see with
the more proscribed string-editing techniques of earlier approaches,
without sacrificing consistency. Third, a subset of the resulting
sentences were sent to a validation task aimed at providing a highly 
reliable set of annotations over the same data, and at identifying areas of inferential uncertainty.





\subsection{Data collection}

\begin{figure}
\begin{framed}
\small
We will show you the caption for a photo. We will not show you the photo. Using only the caption and what you know about the world:
\begin{itemize}
\item Write one alternate caption that is \textbf{definitely} a \textbf{true} description of the photo. \ii{Example: For the caption ``\ii{Two dogs are running through a field.}'' you could write ``\ii{There are animals outdoors.}"}
\item Write one alternate caption that \textbf{might be} a \textbf{true} description of the photo. \ii{Example: For the caption ``\ii{Two dogs are running through a field.}" you could write ``\ii{Some puppies are running to catch a stick.}"}
\item Write one alternate caption that is \textbf{definitely} a \textbf{false} description of the photo. \ii{Example: For the caption ``\ii{Two dogs are running through a field.}" you could write ``\ii{The pets are sitting on a couch.}" This is different from the} maybe correct \ii{category because it's impossible for the dogs to be both running and sitting.}
\end{itemize}
\end{framed}

\caption{\label{instructions-1}The instructions used on Mechanical Turk for data collection.}
\end{figure}

We used Amazon Mechanical Turk for data collection. In each individual task (each HIT), a worker was presented with premise scene descriptions from a pre-existing corpus, and asked to supply hypotheses for each of our three labels---\ii{entailment}, \ii{neutral}, and \ii{contradiction}---forcing the data to be balanced among these classes.

The instructions that we provided to the workers are shown in \reffig{instructions-1}. Below the instructions were three fields for each of three requested sentences, corresponding to our \ii{entailment}, \ii{neutral}, and \ii{contradiction} labels, a fourth field (marked optional) for reporting problems, and a link to an FAQ page. That FAQ grew over the course of data collection. It warned about disallowed techniques (e.g., reusing the same sentence for many different prompts, which we saw in a few cases), provided guidance concerning sentence length and complexity (we did not enforce a minimum length, and we allowed bare NPs as well as full sentences), and reviewed logistical issues around payment timing. About 2,500 workers contributed.

For the premises, we used captions from the Flickr30k corpus \cite{hodoshimage}, a collection of approximately 160k captions (corresponding to about 30k images) collected in an earlier crowdsourced effort.\footnote{
We additionally include about 4k sentence pairs from a pilot study in which the premise sentences were instead drawn from the VisualGenome corpus (under construction; \texttt{\href{https://visualgenome.org/}{visualgenome.org}}). These examples appear only in the training set, and have pair identifiers prefixed with \texttt{vg} in our corpus.
 } 
The captions were not authored by the photographers who took the source images, and they tend to contain relatively literal scene descriptions that are suited to our approach, rather than those typically associated with personal photographs (as in their example: \word{Our trip to the Olympic Peninsula}). In order to ensure that the label for each sentence pair can be recovered solely based on the available text, we did not use the images at all during corpus collection.

\begin{figure}
\center
\includegraphics[width=3.05in]{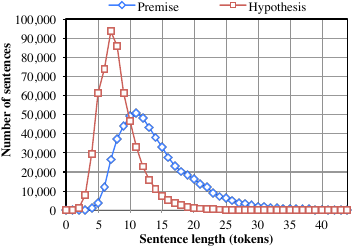}
\caption{\label{length-dist}The distribution of sentence length.} 
\end{figure}

\Reftab{collection-stats} reports some key statistics about the collected corpus, and \reffig{length-dist} shows the distributions of sentence lengths for both our source hypotheses and our newly collected premises. We observed that while premise sentences varied considerably in length, hypothesis sentences tended to be as short as possible while still providing enough information to yield a clear judgment, clustering at around seven words. We also observed that the bulk of the sentences from both sources were syntactically complete rather than fragments, and the frequency with which the parser produces a parse rooted with an `S' (sentence) node attests to this.

\begin{table}
\center
  \begin{tabular}{l r} 
    \toprule
\multicolumn{2}{l}{\textbf{Data set sizes:}}\\
Training pairs &  550,152\\
Development pairs &  10,000\\
Test pairs & 10,000\\
\midrule
\multicolumn{2}{l}{\textbf{Sentence length:}}\\
Premise mean token count & 14.1\\
Hypothesis mean token count & 8.3 \\
\midrule
\multicolumn{2}{l}{\textbf{Parser output:}}\\
Premise `S'-rooted parses & 74.0\%\\
Hypothesis `S'-rooted parses & 88.9\%\\
Distinct words (ignoring case) & 37,026\\
    \bottomrule
  \end{tabular}
\caption{\label{collection-stats}Key statistics for the raw sentence pairs in SNLI\@. Since the two halves of each pair were collected separately, we report some statistics for both.} 
\end{table}

\subsection{Data validation}

In order to measure the quality of our corpus, and in order to construct maximally useful testing and development sets, we performed an additional round of validation for about 10\% of our data.
This validation phase followed the same basic form as the Mechanical Turk labeling task used to label the SICK entailment data: we presented workers with pairs of sentences in batches of five, and asked them to choose a single label for each pair. We supplied each pair to four annotators, yielding five labels per pair including the label used by the original author. The instructions were similar to the instructions for initial data collection shown in \reffig{instructions-1}, and linked to a similar FAQ\@. Though we initially used a very restrictive qualification (based on past approval rate) to select workers for the validation task, we nonetheless discovered (and deleted) some instances of random guessing in an early batch of work, and subsequently instituted a fully closed qualification restricted to about 30 trusted workers.

For each pair that we validated, we assigned a gold label. If any one of
the three labels was chosen by at least three of the five annotators, it was 
chosen as the gold label. If there was no such consensus, which
occurred in about 2\% of cases, we assigned the placeholder label `-'. 
While these unlabeled examples are included in the corpus distribution, they are
unlikely to be helpful for the standard NLI classification task, and
we do not include them in either training or evaluation in the experiments that we 
discuss in this paper.

The results of this validation process
are summarized in \reftab{validation-stats}. 
Nearly all of the examples received a majority
label, indicating broad consensus about the nature of the data and
categories. The gold-labeled examples are very nearly evenly
distributed across the three labels. The Fleiss $\kappa$ scores 
(computed over every example with a full five annotations)
are likely to be conservative given our large and
unevenly distributed pool of annotators, but they still provide insights
about the levels of disagreement across the three semantic
classes. This disagreement likely reflects not just the limitations of
large crowdsourcing efforts but also the uncertainty inherent in naturalistic NLI\@.
Regardless, the overall rate of agreement is extremely high,
suggesting that the corpus is sufficiently high quality to pose a
challenging but realistic machine learning task.

\begin{table}
\center
  \begin{tabular}{l r} 
    \toprule
\multicolumn{2}{l}{\textbf{General:}}\\
Validated pairs & 56,951\\
Pairs w/ unanimous gold label & 58.3\%\\
\midrule
\multicolumn{2}{l}{\textbf{Individual annotator label agreement:}}\\
Individual label $=$ gold label & 89.0\%\\
Individual label $=$ author's label & 85.8\%\\
\midrule
\multicolumn{2}{l}{\textbf{Gold label/author's label agreement:}}\\
Gold label $=$ author's label & 91.2\%\\
Gold label $\ne$ author's label & 6.8\% \\
No gold label (no 3 labels match) & 2.0\%\\
\midrule
\multicolumn{2}{l}{\textbf{Fleiss $\kappa$:}}\\
    \ii{contradiction} & 0.77 \\
    \ii{entailment} & 0.72 \\
    \ii{neutral} & 0.60 \\
    Overall & 0.70 \\
    \bottomrule
  \end{tabular}
\caption{\label{validation-stats}Statistics for the validated pairs. The \ii{author's label} is the label used by the worker who wrote the premise to create the sentence pair. A \ii{gold label} reflects a consensus of three votes from among the author and the four annotators.} 
\end{table}

\subsection{The distributed corpus}

\Reftab{snli-examples} shows a set of randomly chosen validated examples from the development set with their labels. Qualitatively, we find the data that we collected draws fairly extensively on commonsense knowledge, and that hypothesis and premise sentences often differ structurally in significant ways, suggesting that there is room for improvement beyond superficial word alignment models. We also find the sentences that we collected to be largely fluent, correctly spelled English, with a mix of full sentences and caption-style noun phrase fragments, though punctuation and capitalization are often omitted.

The corpus is available under a CreativeCommons
Attribution-ShareAlike license, the same license used for the Flickr30k source captions. It can be downloaded at:\\\href{http://nlp.stanford.edu/projects/snli/}{\texttt{nlp.stanford.edu/projects/snli/}}

\paragraph{Partition} 

We distribute the corpus with a pre-specified train/test/development split. The test and development sets contain 10k examples each. Each original ImageFlickr caption occurs in only one of the three sets, and all of the examples in the test and development sets have been validated.

\paragraph{Parses}

The distributed corpus includes parses produced by the Stanford PCFG Parser 3.5.2 \cite{klein2003accurate}, trained on the standard training set as well as on the Brown Corpus (\citealt{francis1979brown}), which we found to improve the parse quality of the descriptive sentences and noun phrases found in the descriptions. 



\section{Our data as a platform for evaluation}

The most immediate application for our corpus is in developing models for the task of NLI\@. In particular, since it is dramatically larger than any existing corpus of comparable quality, we expect it to be suitable for training parameter-rich models like neural networks, which have not previously been competitive at this task. Our ability to evaluate standard classifier-base NLI models, however, was limited to those which were designed to scale to SNLI's size without modification, so a more complete comparison of approaches will have to wait for future work. In this section, we explore the performance of three classes of models which could scale readily: (i)~models from a well-known NLI system, the Excitement Open Platform; (ii)~variants of a strong but simple feature-based classifier model, which makes use of both unlexicalized and lexicalized features, and (iii)~distributed representation models, including a baseline model and neural network sequence models.

%
%

\subsection{Excitement Open Platform models}

The first class of models is from the Excitement Open
  Platform (EOP,
  \citealt{pado2014design,magnini2014excitement})---an open source platform for RTE research.
%
EOP is a tool for quickly developing NLI systems
  while sharing components such as common lexical resources and 
  evaluation sets.
We evaluate on two algorithms included in the distribution:
  a simple edit-distance based algorithm and
  a classifier-based algorithm, the latter both in a bare form and augmented 
  with EOP's full suite of lexical resources.

%

%
%

\begin{table}
\begin{center}
\def\t#1{\small{#1}}
\begin{tabular}{l@{\hskip \colspaceL}c@{\hskip \colspaceL}c@{\hskip \colspaceL}c}
\toprule
\b{System} & \b{SNLI} & \b{SICK} & \b{RTE-3} \\
\midrule
\t{Edit Distance Based}            & \t{71.9} & \t{65.4} & \t{61.9} \\
\t{Classifier Based}               & \t{72.2} & \t{71.4} & \t{61.5} \\
\t{$~~~$ + Lexical Resources} & \b{75.0} & \b{78.8} & \b{63.6} \\
\bottomrule
\end{tabular}
\end{center}
\caption{
\label{tab:eopresults}
2-class test accuracy for two simple baseline systems included in the
  Excitement Open Platform, as well as SICK and RTE results for
  a model making use of more sophisticated lexical resources.
}
\end{table}
%
%

Our initial goal was to better understand the difficulty of the task of classifying SNLI corpus inferences, rather than necessarily the performance of a state-of-the-art RTE system.
We approached this by running the same system on several data sets:
our own test set,
the SICK test data, and the standard RTE-3 test set \cite{giampiccolo2007third}.
We report results in \reftab{tab:eopresults}.
Each of the models was separately trained on the training set of each corpus.
All models are evaluated only on 2-class entailment.
To convert 3-class problems like SICK and SNLI to this setting, all instances
  of \contradiction\ and \unknown\ are converted to nonentailment.
This yields a most-frequent-class baseline accuracy of 66\% on SNLI, and 71\% on SICK\@.
This is intended primarily to demonstrate the difficulty of the task, rather than necessarily
  the performance of a state-of-the-art RTE system.
The edit distance algorithm tunes the weight of the three 
  case-insensitive edit distance operations on the training set, 
  after removing stop words.
In addition to the base classifier-based system distributed with the platform, we
  train a variant which includes information from
  WordNet \cite{miller1995wordnet} and VerbOcean
  \cite{chklovski2004verbocean}, and makes use of features
  based on tree patterns and dependency tree skeletons
  \cite{wang2007recognizing}.

  


%
%
\subsection{Lexicalized Classifier}
Unlike the RTE datasets, SNLI's size supports approaches which make use of rich lexicalized features.
We evaluate a simple lexicalized classifier to explore the ability of non-specialized models to exploit these features in lieu of more involved language understanding.
Our classifier implements 6 feature types; 3 unlexicalized and 3 lexicalized:
\begin{enumerate}
\setlength\itemsep{-0.25em}
  \item The BLEU score of the \hypothesis\ with respect
  to the \premise, using an n-gram length between 1 and 4.

  \item The length difference between the \hypothesis\ and the \premise, as a real-valued
  feature.

  \item The overlap between words in the \premise\ and \hypothesis,
  both as an absolute count and a percentage of possible overlap, and both over 
  all words and over just nouns, verbs, adjectives, 
  and adverbs.
  
  \item\label{lst:ngram} An indicator for every unigram and bigram in the \hypothesis.

  \item\label{lst:unigram} Cross-unigrams: for every pair of words across the \premise\ and \hypothesis\ which share a 
  POS tag, an indicator feature over the two words.
  
  \item\label{lst:bigram} Cross-bigrams: for every pair of bigrams across the \premise\ and \hypothesis\ which share a 
  POS tag on the second word, an indicator feature over the two bigrams.
\end{enumerate}

%
%

\begin{table}
\begin{center}
\begin{tabular}{l@{\hskip \colspaceL}c@{\hskip \colspaceS}c@{\hskip \colspaceL}c@{\hskip \colspaceS}c}
\toprule
\b{System}	 & \multicolumn{2}{c}{\hspace{-1.2em}\b{SNLI}} & \multicolumn{2}{c}{\b{SICK}}\\
 & \t{Train} & \t{Test} & \t{Train} & \t{Test}\\
\midrule
\t{Lexicalized}            & \t{99.7}  & \b{78.2} & \t{90.4} & \b{77.8} \\ 
\t{Unigrams Only}          & \t{93.1} & \t{71.6}  & \t{88.1} & \t{77.0} \\ 
\t{Unlexicalized}          & \t{49.4} & \t{50.4}  & \t{69.9} & \t{69.6} \\ 
\bottomrule
\end{tabular}
\end{center}
\caption{
\label{tab:bowresults}
3-class accuracy, training on either our data or SICK, including models lacking cross-bigram features 
  (Feature \ref{lst:bigram}), and lacking all lexical
  features (Features \ref{lst:ngram}--\ref{lst:bigram}).
We report results both on the test set and the training set to judge overfitting.
}
\end{table}
%
%

We report results in \reftab{tab:bowresults}, along with ablation studies for removing
  the cross-bigram features (leaving only the cross-unigram feature)
  and for removing all lexicalized features.
On our large corpus in particular, there is a substantial jump in accuracy from using
  lexicalized features, and another from using the very sparse
  cross-bigram features.
The latter  result suggests that there is value in letting
  the classifier automatically learn to recognize structures like explicit negations and adjective
  modification. A similar result was shown in
  \newcite{sidaw12simple} for bigram features in sentiment analysis.
  
It is surprising that the classifier performs as well as it
  does without any notion of alignment or tree transformations.
Although we expect that richer models would perform better,
  the results suggest that given enough data, cross bigrams with the noisy 
  part-of-speech overlap constraint can produce an effective model.


\subsection{Sentence embeddings and NLI}\label{sentence-embedding}

SNLI is suitably large and diverse to make it possible to train neural network models that produce distributed representations of sentence meaning. In this section, we compare the performance of three such models on the corpus. To focus specifically on the strengths of these models at producing informative sentence representations, we use sentence embedding as an intermediate step in the NLI classification task: each model must produce a vector representation of each of the two sentences without using any context from the other sentence, and the two resulting vectors are then passed to a neural network classifier which predicts the label for the pair. This choice allows us to focus on existing models for sentence embedding, and it allows us to evaluate the ability of those models to learn useful representations of meaning (which may be independently useful for subsequent tasks), at the cost of excluding from consideration possible strong neural models for NLI that directly compare the two inputs at the word or phrase level.

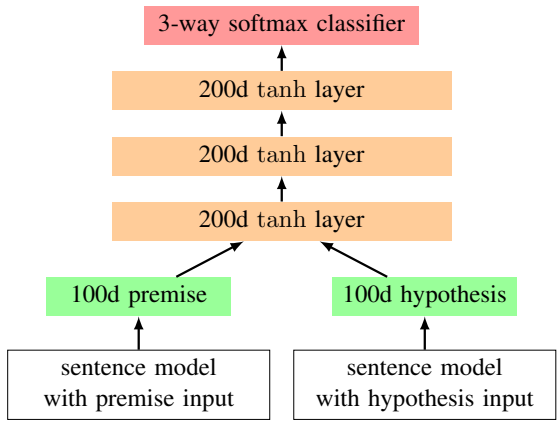
\begin{figure}[tp]
  \centering
\scalebox{0.85}{
 \begin{tikzpicture}
    \def\dx{21pt}
    \def\dy{29pt}

    \tikzstyle{label}=[text width=40mm,align=center]    
    \tikzstyle{softmax}=[fill=red!40,text width=40mm,align=center]
    \tikzstyle{preclass}=[fill=orange!40,text width=50mm,align=center]
    \tikzstyle{e}=[fill=green!40,text width=26mm,align=center]
    \tikzstyle{m}=[draw=black,text width=38mm,align=center]    
    
    \node[softmax]  (softmax) at (0*\dx,6*\dy) {3-way softmax classifier};
    \node[preclass]  (pc3) at (0*\dx,5*\dy) {200d $\tanh$ layer};
    \node[preclass]  (pc2) at (0*\dx,4*\dy) {200d $\tanh$ layer};
    \node[preclass]  (pc1) at (0*\dx,3*\dy) {200d $\tanh$ layer};
    \node[e]  (pe) at (-3*\dx,1.85*\dy) {100d premise};
    \node[e]  (he) at (3*\dx,1.85*\dy) {100d hypothesis};
    \node[m]  (pem) at (-3*\dx,0.5*\dy) {sentence model\\ with premise input};
    \node[m]  (hem) at (3*\dx,0.5*\dy) {sentence model\\ with hypothesis input};    
    
    \pgfsetarrowsend{latex}
    \tikzstyle{fwd} = [draw=black, line width=1pt]

          \draw [fwd] (pc3) -- (softmax);
          \draw [fwd] (pc2) -- (pc3);
          \draw [fwd] (pc1) -- (pc2);
          \draw [fwd] (pe) -- (pc1);
          \draw [fwd] (he) -- (pc1);
          \draw [fwd] (hem) -- (he);
          \draw [fwd] (pem) -- (pe);

  \end{tikzpicture}}
	
        \caption{The neural network classification architecture: for each sentence embedding model evaluated in \reftabs{tab:nnresults}{tab:transferresults}, two identical copies of the model are run with the two sentences as input, and their outputs are used as the two 100d inputs shown here.}
  \label{modelstructure}
\end{figure}

Our neural network classifier, depicted in \reffig{modelstructure} (and based on a one-layer model in \citealt{Bowman:Potts:Manning:2014}), is simply a stack of three 200d $\tanh$ layers, with the bottom layer taking the concatenated sentence representations as input and the top layer feeding a softmax classifier, all trained jointly with the sentence embedding model itself.

We test three sentence embedding models, each set to use 100d phrase and sentence embeddings. Our baseline sentence embedding model simply sums the embeddings of the words in each sentence. In addition, we experiment with two simple sequence embedding models: a plain RNN and an LSTM RNN \cite{hochreiter1997long}.

The word embeddings for all of the models are initialized with the 300d reference GloVe vectors (840B token version, \citealt{pennington2014glove}) and fine-tuned as part of training. In addition, all of the models use an additional $\tanh$ neural network layer to map these 300d embeddings into the lower-dimensional phrase and sentence embedding space. All of the models are randomly initialized using standard techniques and trained using AdaDelta \cite{zeiler2012adadelta} minibatch SGD until performance on the development set stops improving. We applied L2 regularization to all models, manually tuning the strength coefficient $\lambda$ for each, and additionally applied dropout \cite{srivastava2014dropout} to the inputs and outputs of the sentence embedding models (though not to its internal connections) with a fixed dropout rate. All models were implemented in a common framework for this paper, and the implementations will be made available at publication time.

\begin{table}
\begin{center}
\begin{tabular}{l@{\hskip \colspaceL}@{\hskip \colspaceL}c@{\hskip \colspaceL}c}
\toprule
\textbf{Sentence model} & \b{Train}  & \b{Test}\\
\midrule
\t{100d Sum of words}            & \t{79.3} & \t{75.3} \\

\t{100d RNN}            & \t{73.1} & \t{72.2} \\	

\t{100d LSTM RNN}            & \t{84.8} & \b{77.6} \\

\bottomrule
\end{tabular}
\end{center}
\caption{
\label{tab:nnresults}
Accuracy in 3-class classification on our training and test sets for each model.
}
\end{table}

The results are shown in \reftab{tab:nnresults}. The sum of words model performed slightly worse than the fundamentally similar lexicalized classifier---while the sum of words model can use pretrained word embeddings to better handle rare words, it lacks even the rudimentary sensitivity to word order that the lexicalized model's bigram features provide. Of the two RNN models, the LSTM's more robust ability to learn long-term dependencies serves it well, giving it a substantial advantage over the plain RNN, and  resulting in performance that is essentially equivalent to the lexicalized classifier on the test set (LSTM performance near the stopping iteration varies by up to 0.5\% between evaluation steps). While the lexicalized model fits the training set almost perfectly, the gap between train and test set accuracy is relatively small for all three neural network models, suggesting that research into significantly higher capacity versions of these models would be productive.

\subsection{Analysis and discussion}

\Reffig{fig:bowlearncurve} shows a learning curve for the LSTM and the lexicalized and unlexicalized feature-based models. It shows that the large size of the corpus is crucial to both the LSTM and the lexicalized model, and suggests that additional data would yield still better performance for both. In addition, though the LSTM and the lexicalized model show similar performance when trained on the current full corpus, the somewhat steeper slope for the LSTM hints that its ability to learn arbitrarily structured representations of sentence meaning may give it an advantage over the more constrained lexicalized model on still larger datasets.


\Fig{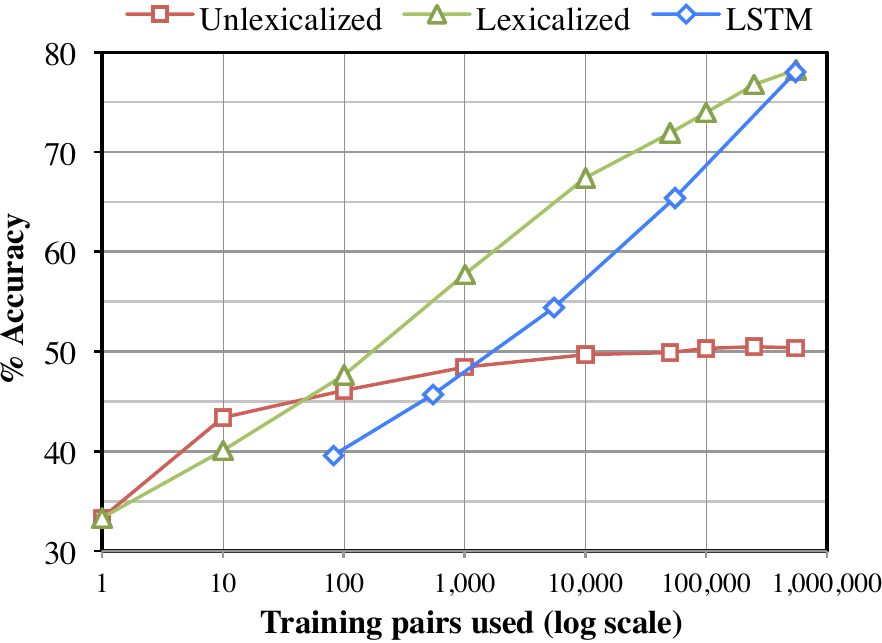}{0.83}{bowlearncurve}{
A learning curve showing how the baseline classifiers and the LSTM perform when trained to convergence on varied amounts of training data. The y-axis starts near a random-chance accuracy of 33\%. The minibatch size of 64 that we used to tune the LSTM sets a lower bound on data for that model.}

We were struck by the speed with which the lexicalized classifier outperforms its unlexicalized counterpart.
With only 100 training examples, the cross-bigram classifier is already performing better.
Empirically, we find that the top weighted features for the classifier
  trained on 100 examples tend to be high precision entailments;
  e.g.,
  \textit{playing} $\rightarrow$ \textit{outside}
  (most scenes are outdoors), \textit{a banana} $\rightarrow$
  \textit{person eating}.
If relatively few spurious entailments get high weight---as it appears
is the case---then it makes sense that, when these do fire, they
boost accuracy in identifying entailments.
  
There are revealing patterns in the errors common to all the models
considered here. Despite the large size of the training corpus and the
distributional information captured by GloVe initialization, many
lexical relationships are still misanalyzed, leading to incorrect
predictions of \ii{independent}, even for pairs that are common in the
training corpus like \word{beach}/\word{surf} and
\word{sprinter}/\word{runner}. Semantic mistakes at the phrasal level
(e.g., predicting contradiction for \word{A male is placing an order in a 
deli}/\word{A man buying a sandwich at a deli}) indicate
that additional attention to compositional semantics would pay off.
%
However, many of the persistent problems run deeper, to inferences
that depend on world knowledge and context-specific inferences, as in
the entailment pair \word{A race car driver leaps from a burning
  car}/\word{A race car driver escaping danger}, for which both
the lexicalized classifier and the LSTM predict \ii{neutral}. 
In other cases, the models' attempts to shortcut this kind of inference 
through lexical cues can lead them astray. 
Some of these examples have qualities
reminiscent of Winograd schemas \cite{Winograd:1972,Levesque:2013}. For
example, all the models wrongly predict
entailment for \word{A young girl throws sand toward the
  ocean}/\word{A girl can't stand the ocean}, presumably because of
distributional associations between \word{throws} and \word{can't
  stand}.

Analysis of the models' predictions also yields insights into the
extent to which they grapple with event and entity coreference. For
the most part, the original image prompts contained a focal element
that the caption writer identified with a syntactic subject, following
information structuring conventions associating subjects and topics in
English \cite{Ward04}. Our annotators generally followed suit, writing
sentences that, while structurally diverse, share topic/focus (theme/rheme)
structure with their premises.
This promotes a coherent, situation-specific construal of each sentence
pair. This is information that our models can easily take advantage
of, but it can lead them astray. For instance, all of them stumble
with the amusingly simple case \emph{A woman prepares ingredients for
  a bowl of soup}/\emph{A soup bowl prepares a woman}, in which prior
expectations about parallelism are not met. Another headline example
of this type is \emph{A man wearing padded arm protection is being
  bitten by a German shepherd dog}/\emph{A man bit a dog}, which all
the models wrongly diagnose as \ii{entailment}, though the sentences
report two very different stories. A model with access
to explicit information about syntactic or semantic structure should perform
better on cases like these.

\section{Transfer learning with SICK}

To the extent that successfully training a neural network model like our LSTM on SNLI forces that model to encode broadly accurate representations of English scene descriptions and to build an entailment classifier over those relations, we should expect it to be readily possible to adapt the trained model for use on other NLI tasks. In this section, we evaluate on the SICK entailment task using a simple transfer learning method \cite{pratt1991direct} and achieve competitive results.

\begin{table}
\begin{center}
\begin{tabular}{l@{\hskip \colspaceL}@{\hskip \colspaceL}r@{\hskip \colspaceL}r}
\toprule
\textbf{Training sets} & \b{Train}  & \b{Test}\\
\midrule
\t{Our data only}            & \t{42.0} & \t{46.7} \\
\t{SICK only}            & \t{100.0} & \t{71.3} \\
\t{Our data and SICK (transfer)}            & \t{99.9} & \b{80.8} \\
\bottomrule
\end{tabular}
\end{center}

\caption{\label{tab:transferresults}
LSTM 3-class accuracy on the SICK train and test sets under three training regimes.} 
\end{table}

To perform transfer, we take the parameters of the LSTM RNN model trained on SNLI and use them to initialize a new model, which is trained from that point only on the training portion of SICK. The only newly initialized parameters are softmax layer parameters and the embeddings for words that appear in SICK, but not in SNLI (which are populated with GloVe embeddings as above). We use the same model hyperparameters that were used to train the original model, with the exception of the L2 regularization strength, which is re-tuned. We additionally transfer the accumulators that are used by AdaDelta to set the learning rates. This lowers the starting learning rates, and is intended to ensure that the model does not learn too quickly in its first few epochs after transfer and destroy the knowledge accumulated in the pre-transfer phase of training.

The results are shown in \reftab{tab:transferresults}. Training on SICK alone yields poor performance, and the model trained on SNLI fails when tested on SICK data, labeling more \ii{neutral} examples as \ii{contradiction}s than correctly, possibly as a result of subtle differences in how the labeling task was presented. In contrast, transferring SNLI representations to SICK yields the best performance yet reported for an unaugmented neural network model, surpasses the available EOP models, and approaches both the overall state of the art at 84.6\% \cite{lai2014illinois} and the 84\% level of interannotator agreement, which likely represents an approximate performance ceiling. This suggests that the introduction of a large high-quality corpus makes it  possible to train representation-learning models for sentence meaning that are competitive with the best hand-engineered models on inference tasks.

We attempted to apply this same transfer evaluation technique to the RTE-3 challenge, but found that the small training set (800 examples) did not allow the model to adapt to the unfamiliar genre of text used in that corpus, such that no training configuration yielded competitive performance.
Further research on effective transfer learning on small data sets with neural models might facilitate improvements here.



\section{Conclusion}\label{sec:conclusion}

Natural languages are powerful vehicles for reasoning, 
and nearly all questions about meaningfulness
in language can be reduced to questions of entailment
and contradiction in context. This suggests that NLI is an ideal testing ground
for theories of semantic representation, and that training for NLI
tasks can provide rich domain-general semantic representations.  To
date, however, it has not been possible to fully realize this
potential due to the limited nature of existing NLI resources.  This
paper sought to remedy this with a new, large-scale, naturalistic
corpus of sentence pairs labeled for entailment, contradiction, and
independence. We used this corpus to evaluate a range of models,
and found that both simple lexicalized models and neural network
models perform well, and that the representations learned by
a neural network model on our corpus can be used to dramatically 
improve performance on a standard challenge dataset. We hope that
SNLI presents valuable training data and a challenging testbed for the continued 
application of machine learning to semantic representation.

\subsubsection*{Acknowledgments}

We gratefully acknowledge support from %
a Google Faculty Research Award, %
a gift from Bloomberg L.P., 
the Defense Advanced Research Projects Agency (DARPA) Deep Exploration and Filtering of Text (DEFT) Program under Air Force Research Laboratory (AFRL) contract no.~FA8750-13-2-0040,
the National Science Foundation under grant no.~IIS 1159679, and %
the Department of the Navy, Office of Naval Research, under grant no.~N00014-10-1-0109.
Any opinions, findings, and conclusions or recommendations expressed in this material are those of the authors and do not necessarily reflect the views of 
Google, 
Bloomberg L.P.,
DARPA,
AFRL
NSF, 
ONR, or 
the US government. We also thank our many excellent Mechanical Turk contributors.


\bibliographystyle{acl}
\bibliography{MLSemantics} 

\begin{thebibliography}{}

\bibitem[\protect\citename{Bos and Markert}2005]{BosMar:2005}
Johan Bos and Katja Markert.
\newblock 2005.
\newblock Recognising textual entailment with logical inference.
\newblock In {\em Proc. EMNLP}.

\bibitem[\protect\citename{Bowman \bgroup et al.\egroup
  }2015]{Bowman:Potts:Manning:2014}
Samuel~R. Bowman, Christopher Potts, and Christopher~D. Manning.
\newblock 2015.
\newblock Recursive neural networks can learn logical semantics.
\newblock In {\em Proc. of the 3rd Workshop on Continuous Vector Space Models
  and their Compositionality}.

\bibitem[\protect\citename{Chklovski and Pantel}2004]{chklovski2004verbocean}
Timothy Chklovski and Patrick Pantel.
\newblock 2004.
\newblock Verb{O}cean: Mining the web for fine-grained semantic verb relations.
\newblock In {\em Proc. {EMNLP}}.

\bibitem[\protect\citename{Condoravdi \bgroup et al.\egroup
  }2003]{Condoravdi-etAl:2003}
Cleo Condoravdi, Dick Crouch, Valeria de~Paiva, Reinhard Stolle, and Daniel~G.
  Bobrow.
\newblock 2003.
\newblock Entailment, intensionality and text understanding.
\newblock In {\em Proc. of the {HLT-NAACL} 2003 Workshop on Text Meaning}.

\bibitem[\protect\citename{Dagan \bgroup et al.\egroup }2006]{dagan2006pascal}
Ido Dagan, Oren Glickman, and Bernardo Magnini.
\newblock 2006.
\newblock The {PASCAL} recognising textual entailment challenge.
\newblock In {\em Machine learning challenges. Evaluating predictive
  uncertainty, visual object classification, and recognising tectual
  entailment}, pages 177--190. Springer.

\bibitem[\protect\citename{de Marneffe \bgroup et al.\egroup
  }2008]{de2008finding}
Marie-Catherine de~Marneffe, Anna~N. Rafferty, and Christopher~D. Manning.
\newblock 2008.
\newblock Finding contradictions in text.
\newblock In {\em Proc. ACL}.

\bibitem[\protect\citename{Francis and Kucera}1979]{francis1979brown}
W.~Nelson Francis and Henry Kucera.
\newblock 1979.
\newblock Brown corpus manual.
\newblock Brown University.

\bibitem[\protect\citename{Fyodorov \bgroup et al.\egroup
  }2000]{Fyodorov-etal:2000}
Yaroslav Fyodorov, Yoad Winter, and Nissim Francez.
\newblock 2000.
\newblock A natural logic inference system.
\newblock In {\em Proc. of the 2nd Workshop on Inference in Computational
  Semantics}.

\bibitem[\protect\citename{Giampiccolo \bgroup et al.\egroup
  }2007]{giampiccolo2007third}
Danilo Giampiccolo, Bernardo Magnini, Ido Dagan, and Bill Dolan.
\newblock 2007.
\newblock The third {PASCAL} recognizing textual entailment challenge.
\newblock In {\em Proc. of the ACL-PASCAL workshop on textual entailment and
  paraphrasing}.

\bibitem[\protect\citename{Hochreiter and Schmidhuber}1997]{hochreiter1997long}
Sepp Hochreiter and J{\"u}rgen Schmidhuber.
\newblock 1997.
\newblock Long short-term memory.
\newblock {\em Neural computation}, 9(8):1735--1780.

\bibitem[\protect\citename{Katz}1972]{Katz72}
Jerrold~J. Katz.
\newblock 1972.
\newblock {\em Semantic Theory}.
\newblock Harper \& Row, New York.

\bibitem[\protect\citename{Klein and Manning}2003]{klein2003accurate}
Dan Klein and Christopher~D. Manning.
\newblock 2003.
\newblock Accurate unlexicalized parsing.
\newblock In {\em Proc. ACL}.

\bibitem[\protect\citename{Lai and Hockenmaier}2014]{lai2014illinois}
Alice Lai and Julia Hockenmaier.
\newblock 2014.
\newblock Illinois-{LH}: A denotational and distributional approach to
  semantics.
\newblock In {\em Proc. {SemEval}}.

\bibitem[\protect\citename{Levesque}2013]{Levesque:2013}
Hector~J. Levesque.
\newblock 2013.
\newblock On our best behaviour.
\newblock In {\em Proc. {AAAI}}.

\bibitem[\protect\citename{Levy \bgroup et al.\egroup }2014]{levy2014focused}
Omer Levy, Ido Dagan, and Jacob Goldberger.
\newblock 2014.
\newblock Focused entailment graphs for open {IE} propositions.
\newblock In {\em Proc. {CoNLL}}.

\bibitem[\protect\citename{MacCartney and Manning}2009]{maccartney2009extended}
Bill MacCartney and Christopher~D Manning.
\newblock 2009.
\newblock An extended model of natural logic.
\newblock In {\em Proc. of the Eighth International Conference on Computational
  Semantics}.

\bibitem[\protect\citename{Magnini \bgroup et al.\egroup
  }2014]{magnini2014excitement}
Bernardo Magnini, Roberto Zanoli, Ido Dagan, Kathrin Eichler, G{\"u}nter
  Neumann, Tae-Gil Noh, Sebastian Pado, Asher Stern, and Omer Levy.
\newblock 2014.
\newblock The {E}xcitement {O}pen {P}latform for textual inferences.
\newblock {\em Proc. ACL}.

\bibitem[\protect\citename{Marelli \bgroup et al.\egroup
  }2014a]{marelli2014semeval}
Marco Marelli, Luisa Bentivogli, Marco Baroni, Raffaella Bernardi, Stefano
  Menini, and Roberto Zamparelli.
\newblock 2014a.
\newblock Sem{E}val-2014 task 1: Evaluation of compositional distributional
  semantic models on full sentences through semantic relatedness and textual
  entailment.
\newblock In {\em Proc. SemEval}.

\bibitem[\protect\citename{Marelli \bgroup et al.\egroup
  }2014b]{marelli2014sick}
Marco Marelli, Stefano Menini, Marco Baroni, Luisa Bentivogli, Raffaella
  Bernardi, and Roberto Zamparelli.
\newblock 2014b.
\newblock A {SICK} cure for the evaluation of compositional distributional
  semantic models.
\newblock In {\em Proc. LREC}.

\bibitem[\protect\citename{Miller}1995]{miller1995wordnet}
George~A Miller.
\newblock 1995.
\newblock {W}ord{N}et: a lexical database for english.
\newblock {\em Communications of the ACM}, 38(11):39--41.

\bibitem[\protect\citename{Pad{\'o} \bgroup et al.\egroup
  }2014]{pado2014design}
Sebastian Pad{\'o}, Tae-Gil Noh, Asher Stern, Rui Wang, and Roberto Zanoli.
\newblock 2014.
\newblock Design and realization of a modular architecture for textual
  entailment.
\newblock {\em Journal of Natural Language Engineering}.

\bibitem[\protect\citename{Pavlick \bgroup et al.\egroup
  }2015]{ganitkevitch2ppdb}
Ellie Pavlick, Johan Bos, Malvina Nissim, Charley Beller, Ben~Van Durme, and
  Chris Callison-Burch.
\newblock 2015.
\newblock {PPDB} 2.0: Better paraphrase ranking, fine-grained entailment
  relations, word embeddings, and style classification.
\newblock In {\em Proc. {ACL}}.

\bibitem[\protect\citename{Pennington \bgroup et al.\egroup
  }2014]{pennington2014glove}
Jeffrey Pennington, Richard Socher, and Christopher~D Manning.
\newblock 2014.
\newblock Glo{V}e: Global vectors for word representation.
\newblock In {\em Proc. {EMNLP}}.

\bibitem[\protect\citename{Pratt \bgroup et al.\egroup }1991]{pratt1991direct}
Lorien~Y Pratt, Jack Mostow, Candace~A Kamm, and Ace~A Kamm.
\newblock 1991.
\newblock Direct transfer of learned information among neural networks.
\newblock In {\em Proc. {AAAI}}.

\bibitem[\protect\citename{Socher \bgroup et al.\egroup
  }2013]{socher2013recursive}
Richard Socher, Alex Perelygin, Jean~Y Wu, Jason Chuang, Christopher~D Manning,
  Andrew~Y Ng, and Christopher Potts.
\newblock 2013.
\newblock Recursive deep models for semantic compositionality over a sentiment
  treebank.
\newblock In {\em Proc. EMNLP}.

\bibitem[\protect\citename{Srivastava \bgroup et al.\egroup
  }2014]{srivastava2014dropout}
Nitish Srivastava, Geoffrey Hinton, Alex Krizhevsky, Ilya Sutskever, and Ruslan
  Salakhutdinov.
\newblock 2014.
\newblock Dropout: {A} simple way to prevent neural networks from overfitting.
\newblock {\em JMLR}.

\bibitem[\protect\citename{van Benthem}2008]{vanBenthem08NATLOG}
Johan van Benthem.
\newblock 2008.
\newblock A brief history of natural logic.
\newblock In M.~Chakraborty, B.~L{\"o}we, M.~Nath~Mitra, and S.~Sarukki,
  editors, {\em Logic, {N}avya-{N}yaya and Applications: Homage to {B}imal
  {M}atilal}. College Publications.

\bibitem[\protect\citename{Wang and Manning}2012]{sidaw12simple}
Sida~I. Wang and Christopher~D. Manning.
\newblock 2012.
\newblock Baselines and bigrams: Simple, good sentiment and topic
  classification.
\newblock In {\em Proc. ACL}.

\bibitem[\protect\citename{Wang and Neumann}2007]{wang2007recognizing}
Rui Wang and G{\"u}nter Neumann.
\newblock 2007.
\newblock Recognizing textual entailment using sentence similarity based on
  dependency tree skeletons.
\newblock In {\em ACL-PASCAL Workshop on Textual Entailment and Paraphrasing}.

\bibitem[\protect\citename{Ward and Birner}2004]{Ward04}
Gregory Ward and Betty Birner.
\newblock 2004.
\newblock Information structure and non-canonical syntax.
\newblock In Laurence~R. Horn and Gregory Ward, editors, {\em Handbook of
  Pragmatics}, pages 153--174. Blackwell, Oxford.

\bibitem[\protect\citename{Weston \bgroup et al.\egroup
  }2015a]{Weston-etal:2015Q}
Jason Weston, Antoine Bordes, Sumit Chopra, and Tomas Mikolov.
\newblock 2015a.
\newblock Towards {AI}-complete question answering: A set of prerequisite toy
  tasks.
\newblock arXiv:1502.05698.

\bibitem[\protect\citename{Weston \bgroup et al.\egroup
  }2015b]{Weston-etal:2015}
Jason Weston, Sumit Chopra, and Antoine Bordes.
\newblock 2015b.
\newblock Memory networks.
\newblock In {\em Proc. {ICLR}}.

\bibitem[\protect\citename{Winograd}1972]{Winograd:1972}
Terry Winograd.
\newblock 1972.
\newblock Understanding natural language.
\newblock {\em Cognitive Psychology}, 3(1):1--191.

\bibitem[\protect\citename{Young \bgroup et al.\egroup }2014]{hodoshimage}
Peter Young, Alice Lai, Micah Hodosh, and Julia Hockenmaier.
\newblock 2014.
\newblock From image descriptions to visual denotations: New similarity metrics
  for semantic inference over event descriptions.
\newblock {\em TACL}, 2:67--78.

\bibitem[\protect\citename{Zeiler}2012]{zeiler2012adadelta}
Matthew~D. Zeiler.
\newblock 2012.
\newblock {ADADELTA}: an adaptive learning rate method.
\newblock {\em arXiv preprint arXiv:1212.5701}.

\end{thebibliography}

\end{document}